\documentclass[pre, a4paper, superscriptaddress, amsfonts, amssymb, amsmath, reprint, showkeys, nofootinbib, oneside]{revtex4-1}
\usepackage[english]{babel}
\usepackage[utf8]{inputenc}
\usepackage{xcolor}
\usepackage{bm}% bold math
\usepackage{graphicx}% Include figure files
\usepackage{float}
\usepackage{amsmath}
\usepackage{amsfonts,amssymb,amsthm}
\usepackage[colorinlistoftodos, color=green!40, prependcaption]{todonotes}
\usepackage{amsthm}
\usepackage{mathtools}
\usepackage{physics}
\usepackage{graphicx}
\usepackage[left=23mm,right=13mm,top=35mm,columnsep=15pt]{geometry} 
\usepackage{adjustbox}
\usepackage{placeins}
\usepackage[T1]{fontenc}
\usepackage{lipsum}
\usepackage{csquotes}
\usepackage[pdftex, pdftitle={Article}, pdfauthor={Author}]{hyperref} % For hyperlinks in the PDF
\usepackage{notes2bib}
\usepackage{hyperref}
\usepackage{filecontents}
\usepackage{xcolor}

\DeclareMathOperator*{\argminA}{arg\,min} % Jan Hlavacek
\DeclareMathOperator*{\argmaxA}{arg\,max}

\begin{document}
\title{Information-theoretic Quantification of High-order Feature Effects \\in Classification Problems}

\author{Ivan Lazic}
    \affiliation{Faculty of Technical Sciences, University of Novi Sad, Novi Sad, Serbia}   
\author{Chiara Barà}
    \affiliation{Department of Engineering, University of Palermo, Palermo, Italy}
\author{Marta Iovino}
    \affiliation{Department of Engineering, University of Palermo, Palermo, Italy}
\author{Sebastiano Stramaglia}
    \affiliation{Dipartimento Interateneo di Fisica, Università degli Studi di Bari Aldo Moro, Bari, Italy}
\author{Niksa Jakovljevic}
    \affiliation{Faculty of Technical Sciences, University of Novi Sad, Novi Sad, Serbia}       
\author{Luca Faes}
    \email[Correspondence email address: ]{luca.faes@unipa.it}% Your name
    \affiliation{Department of Engineering, University of Palermo, Palermo, Italy}
    \affiliation{Faculty of Technical Sciences, University of Novi Sad, Novi Sad, Serbia} 

\begin{abstract}
    Understanding the contribution of individual features in predictive models remains a central goal in interpretable machine learning, and while many model-agnostic methods exist to estimate feature importance, they often fall short in capturing high-order interactions and disentangling overlapping contributions. In this work, we present an information-theoretic extension of the High-order interactions for Feature importance (Hi-Fi) method, leveraging Conditional Mutual Information (CMI) estimated via a k-Nearest Neighbor (kNN) approach working on mixed discrete and continuous random variables. Our framework decomposes feature contributions into unique, synergistic, and redundant components, offering a richer, model-independent understanding of their predictive roles. We validate the method using synthetic datasets with known Gaussian structures, where ground truth interaction patterns are analytically derived, and further test it on non-Gaussian and real-world gene expression data from TCGA-BRCA. Results indicate that the proposed estimator accurately recovers theoretical and expected findings, providing a potential use case for developing feature selection algorithms or model development based on interaction analysis.
\end{abstract}

%\keywords{first keyword, second keyword, third keyword}

\maketitle

\section{Introduction}

One of the central aspects of model explainability in machine learning is the identification and interpretation of feature importance \cite{Arsenault_2025}. Understanding which features drive the predictions, on both local and global levels, not only enhances the user's trust in the automated decision process but also provides valuable insights into the underlying data structure and feature interactions within the prediction process. As a consequence, numerous methods have been proposed for assessing feature contributions in a model-agnostic setting, ranging from simple error-based permutation importance tests \cite{breiman2001random} and basic information-theoretic approaches \cite{Vergara_2013}, to more sophisticated techniques such as SHAP, which relies on the weighted average contribution of features across possible subsets \cite{lundberg2017unifiedapproachinterpretingmodel}.

Despite their success, these methods often fail to isolate true marginal contributions of individual features, particularly in the presence of high-order interactions, cases where the importance of a feature depends not only on its contribution but also on its combined influence with other features. Such interactions can make it difficult to determine whether a feature provides unique information alone or in conjunction with specific feature subsets, leading to synergistic or redundant effects. To address this, König \textit{et al.} \cite{könig2024disentanglinginteractionsdependenciesfeature} propose a model-agnostic approach to disentangle standalone, interaction, and dependency effects by decomposing the explained variance of a pretrained model's prediction. In contrast, Ontivero-Ortega \textit{et al.} \cite{ontivero2025assessing} leverage the concept of Partial Information Decomposition (PID) \cite{williams2010nonnegativedecompositionmultivariateinformation} and build upon the Leave-One-Covariate-Out (LOCO) approach \cite{Lei03072018} to resolve the prediction error reduction in a regression problem into unique, synergistic, and redundant contributions of the features. The crucial novelty of this approach is that it allows to quantify the High-order effects in Feature importance (Hi-Fi method), capturing both the marginal effect of excluding a feature and its interactions with subsets of other features in contributing to the model's output.
% Ontivero-Ortega \textit{et al.} \cite{ontivero2025assessing} build upon the Leave-One-Covariate-Out (LOCO) approach \cite{Lei03072018}, leveraging the framework of the Partial Information Decomposition (PID) \cite{williams2010nonnegativedecompositionmultivariateinformation}, to resolve the prediction error reduction in a regression problem into unique, synergistic, and redundant contributions of the features. This approach allows to quantify the high-order effects in feature importance (Hi-Fi method). This extension captures the marginal effect of excluding a feature while also revealing its interaction with subsets of other features in contributing to the model's output. 
% As noted in \cite{ontivero2025assessing}, based on the work by Stramaglia \textit{et al.} on the decomposition of Granger Causal effects \cite{stramaglia2024disentangling}, the change in predictive performance induced by omitting a feature can be interpreted as a proxy for the Conditional Mutual Information (CMI) between the target and the source variable, conditioned on the remaining features.

In this work, we build upon this foundation to bring the Hi-Fi method into the context of information theory, which enables the decomposition of feature contributions into interpretable informational components, providing a rich, model-independent characterization of feature importance beyond marginal effects. As noted in \cite{ontivero2025assessing}, based on the work by Stramaglia \textit{et al.} on the decomposition of Granger causal effects \cite{stramaglia2024disentangling}, the change in predictive performance induced by omitting a feature can be interpreted as a proxy for the Conditional Mutual Information (CMI) between the target and the source variable, conditioned on the remaining features. Furthermore, to make the approach computationally reliable in practical classification problems, we design a data-efficient estimator of the CMI, based on the k-Nearest Neighbor (kNN) method, computed between a discrete output class variable and several continuous input feature variables. We validate the approach utilizing simulated data with known ground truth values, as well as simulated and real-world data without ground truth, but with expected behavior derived from controlled interaction setups or supported by existing literature.

\section{Method}

Let us consider a system with $n+2$ variables, out of which one represents the class variable $Y$, while the rest represent continuous features. For the analysis, we isolate the variable $X$, denoted as the source feature, for which we assess the importance in relation to the other features collected in  $\boldsymbol{Z}=\{Z_1,\dots,Z_n\}$. Our goal is to determine the importance of $X$ to explain $Y$ in the context of the variables $\boldsymbol{Z}$. To this end, we introduce an information-theoretic equivalent of the LOCO approach \cite{Lei03072018} exploiting the CMI between the class and the source feature conditioned to the remaining features, i.e.
\begin{equation}  \label{eq:IYX_Z}
    I(Y;X|\boldsymbol{Z}) = I(Y;X,\boldsymbol{Z}) - I(Y; \boldsymbol{Z}).
\end{equation}
Intuitively, the CMI (1) captures the information provided to the target $Y$ by the source $X$ above and beyond the information brought by the other sources in $\boldsymbol{Z}$. This measure was  investigated in \cite{wollstadt2023rigorousinformationtheoreticdefinitionredundancy} in terms of PID, showing that it collects the unique and synergistic effects obtained from the source feature, providing a well-principled criterion for feature selection.
Here, using the searching approach devised in \cite{ontivero2025assessing}, we take another perspective on higher-order effects by looking for sets of variables in $\boldsymbol{Z}$ that minimize or maximize the CMI between the class $Y$ and the source $X$, i.e., $I(Y;X|\boldsymbol{Z}_{min})$ and $I(Y;X|\boldsymbol{Z}_{max})$. In this way we disentangle the unique contribution of $X$ in determining $Y$ from that deriving from its redundant and synergistic interaction with the variables in $\boldsymbol{Z}$ without the need to perform PID. Specifically, the maximum information shared between the class $Y$ and the source $X$ can be decomposed as follows \cite{stramaglia2024disentangling}:
\begin{equation}  \label{eq:IYX_Zmax}
\begin{split}
    I(Y;X|\boldsymbol{Z}_{max}) &= S(Y;X|\boldsymbol{Z})+R(Y;X|\boldsymbol{Z})  +U(Y;X|\boldsymbol{Z}),
\end{split}
\end{equation}
with 
\begin{align}
\label{eq:URS}
\begin{split}
 S(Y;X|\boldsymbol{Z}) &= I(Y;X|\boldsymbol{Z}_{max}) - I(Y;X) ,
\\
 R(Y;X|\boldsymbol{Z}) &= I(Y;X) - I(Y;X|\boldsymbol{Z}_{min}) ,
 \\
 U(Y;X|\boldsymbol{Z}) &= I(Y;X|\boldsymbol{Z}_{min}) ,
\end{split}
\end{align}
representing the synergistic, redundant, and unique contribution, respectively.

% $S(Y;X|\boldsymbol{Z}) = I(Y;X|\boldsymbol{Z}_{max}) - I(Y;X)$ the synergistic contribution, $R(Y;X|\boldsymbol{Z}) = I(Y;X) - I(Y;X|\boldsymbol{Z}_{min})$ the redundant contribution, and $U(Y;X|\boldsymbol{Z}) = I(Y;X|\boldsymbol{Z}_{min})$ the unique contribution.

\subsection{Searching algorithm} \label{searchAlg}
Due to the computational complexity of an exhaustive search, a greedy approach is typically preferred to obtain $\boldsymbol{Z}_{min}$ and $\boldsymbol{Z}_{max}$. In detail, for any given iteration of a sequential procedure, variables are added to the two subsets one at a time, minimizing or maximizing the CMI quantities with respect to the set of previously selected variables. %$\boldsymbol{Z}_{j-1}$.
The procedure is initialized representing the subsets of selected variables as the empty set, $\boldsymbol{Z}_0 = \emptyset$.
%, and computing the corresponding CMI $I(Y;X|\boldsymbol{Z}_0) = I(Y; X)$, which is simply the Mutual Information (MI) between the source variable $X$ and the class $Y$.
% This value is set as the initial $I(Y;X|\boldsymbol{Z}_{min})$ and $I(Y;X|\boldsymbol{Z}_{max})$ that is iteratively updated depending on the collected conditioning. 
Then, to determine the set $\boldsymbol{Z}_{min}$, at any given step $j$ ($1\leq j \leq n$) the variable $V_j \in \boldsymbol{Z} \setminus \boldsymbol{Z}_{j-1}$ is determined such that
%a search in individual variables $V \in \boldsymbol{Z} \setminus \boldsymbol{Z}_{j-1}$ is performed that minimizes the CMI, i.e.,
\begin{equation}
    j = \argminA_{V \in \boldsymbol{Z} \setminus  \boldsymbol{Z}_{j-1}} I(Y;X|\boldsymbol{Z}_{j-1},V).
    %I(Y;X|\boldsymbol{Z}_{j-1},V_i) = \min_{V \in \boldsymbol{Z} \setminus  \boldsymbol{Z}_{j-1}} I(Y;X|\boldsymbol{Z}_{j-1},V).
\end{equation}
If the CMI $I(Y;X|\boldsymbol{Z}_{j-1},V_j)$  decreases significantly below $I(Y;X|\boldsymbol{Z}_{j-1})$, the variable $V_j$ is retained, the set of selected variables is updated as $\boldsymbol{Z}_j = \{\boldsymbol{Z}_{j-1}, V_j\}$ and the counter $j$ is updated; otherwise, if the decrease in CMI is not significant, $V_j$ is discarded and the search terminates with $\boldsymbol{Z}_{min}=\boldsymbol{Z}_{j-1}$.
%there is no significant decrease in the CMI quantity for an observed $V_j$. The final selected set represents the $\boldsymbol{Z}_{min}$.
Similarly, to determine the set $\boldsymbol{Z}_{max}$ which maximizes the CMI starting again from the empty set $\boldsymbol{Z}_0 = \emptyset$, at each step $j \geq 1$ the variable $V_j \in \boldsymbol{Z} \setminus \boldsymbol{Z}_{j-1}$ is determined such that 
\begin{equation}
    j = \argmaxA_{V \in \boldsymbol{Z} \setminus  \boldsymbol{Z}_{j-1}} I(Y;X|\boldsymbol{Z}_{j-1},V);
\end{equation}
$V_j$ is retained updating the set of selected features as $\boldsymbol{Z}_j = \{\boldsymbol{Z}_{j-1}, V_j\}$ if the increase of the CMI $I(Y;X|\boldsymbol{Z}_{j-1},V_j)$ above $I(Y;X|\boldsymbol{Z}_{j-1})$ is significant, or it is otherwise discarded terminating the search with $\boldsymbol{Z}_{max}=\boldsymbol{Z}_{j-1}$.
After the two searches, the unique, redundant and synergistic contributions can be determined applying \eqref{eq:URS} with the selected $\boldsymbol{Z}_{min}$ and $\boldsymbol{Z}_{max}$.
% , until the corresponding decrease or increase is statistically significant according to surrogate analysis.

\subsection{Estimation approach}
To perform the searching algorithm and find $\boldsymbol{Z}_{min}$ and $\boldsymbol{Z}_{max}$, at the generic step $j$ we need to compute the two CMI terms to be compared for quantifying the variation in the predictive improvement brought by the candidate variable $V$, i.e.,
\begin{align} 
\label{eq:CMI_formule}
\begin{split}
    I(Y;X|\boldsymbol{Z}_{j-1},V) &= I(Y;X,\boldsymbol{Z}_{j-1},V)    -I(Y;\boldsymbol{Z}_{j-1},V),
    \\
    I(Y;X|\boldsymbol{Z}_{j-1}) &= I(Y;X,\boldsymbol{Z}_{j-1})  - I(Y;\boldsymbol{Z}_{j-1}).
\end{split}
\end{align}
In the following, we first describe the estimator developed in this work to compute the two CMI values in (\ref{eq:CMI_formule}) from a dataset of $N$ observations of the class and feature variables, and then describe the procedure followed to assess the statistical significance of the CMI variation obtained adding the selected $V_j$ to $\boldsymbol{Z}_{j-1}$ inside the conditioning set.

The two CMI values are estimated using a mixed variable approach based on the kNN estimator of the mutual information \cite{kraskov2004estimating}.
%which accounts for the discrete nature of the class variable $Y$ and the continuous nature of the features $X$ and $\boldsymbol{Z}$.
Specifically, the mixed approach considers the class variable $Y$ as discrete, with alphabet $\mathcal{A}_Y$, and the feature variables $X$ and $\boldsymbol{Z}$ as continuous, with domains $\mathcal{D}_X$ and $\mathcal{D}_{\boldsymbol{Z}}$, respectively. Then, a mixed kNN estimation approach is followed \cite{bara2024partial} to compute the four MI terms in \eqref{eq:CMI_formule}, adapting the kNN MI estimator \cite{kraskov2004estimating} to consider the estimation bias possibly arising with the comparison across different neighborhood spaces \cite{PhysRevE.95.062114}. In order to minimize such bias, a neighbor search is performed first in the highest-dimensional space $\{X,\boldsymbol{Z}_{j-1}, V\}$, after which the determined distances are projected to the lower-dimensional spaces $\{\boldsymbol{Z}_{j-1}, V\}$, $\{X, \boldsymbol{Z}_{j-1}\}$, and $\{\boldsymbol{Z}_{j-1}\}$.
In detail, each MI term is calculated as the weighted sum of the specific MI relevant to the outcome $y$ of the class $Y$, where the weights are given by the probability of the event $Y=y$, i.e., $P(\{Y=y\})=P(y)$. The specific MI terms are computed as:
\begin{align}
I(Y=y;X,\boldsymbol{Z}_{j-1},V) &= \psi(N) - \psi(N_y) + \psi(k)  \nonumber\\&-\langle\psi(m_{x,z_{j-1},v}+1)\rangle, \\
I(Y=y;\boldsymbol{Z}_{j-1},V) &= \psi(N) - \psi(N_y) +\langle\psi(m_{z_{j-1},v}^y+1)\rangle \nonumber\\
    &\quad -\langle\psi(m_{z_{j-1},v}+1)\rangle,  \\
I(Y=y;X,\boldsymbol{Z}_{j-1}) &= \psi(N) - \psi(N_y) \label{eq:IyX_Z} +\langle\psi(m_{x,z_{j-1}}^y+1)\rangle \nonumber \\ 
    &\quad -\langle\psi(m_{x,z_{j-1}}+1)\rangle,\\
I(Y=y;\boldsymbol{Z}_{j-1}) &= \psi(N) - \psi(N_y) \label{eq:Iy_Z} +\langle\psi(m_{z_{j-1}}^y+1)\rangle  \nonumber \\ 
    &\quad-\langle\psi(m_{z_{j-1}}+1)\rangle,
\end{align}
% \begin{align}
%     &I(Y=y;X,\boldsymbol{Z}_{j-1},V) = \psi(N) - \psi(N_y) + \psi(k) \\ 
%     &-\frac{1}{N_y}\sum_{\{(x,z_{j-1},v)^y\}}\psi(m_{x,z_{j-1},v}+1), \nonumber\\
%     &I(Y=y;\boldsymbol{Z}_{j-1},V) = \psi(N) - \psi(N_y) \\
%     &+\frac{1}{N_y}\sum_{\{(z_{j-1},v)^y\}}(\psi(m_{z_{j-1},v}^y+1) - \psi(m_{z_{j-1},v}+1)),\nonumber\\
%    & I(Y=y;X,\boldsymbol{Z}_{j-1}) = \psi(N) - \psi(N_y) \label{eq:IyX_Z}  \\ 
%     &+\frac{1}{N_y}\sum_{\{(x,z_{j-1})^y\}}(\psi(m_{x,z_{j-1}}^y+1) - \psi(m_{x,z_{j-1}}+1)),\nonumber\\
%     &I(Y=y;\boldsymbol{Z}_{j-1}) = \psi(N) - \psi(N_y) \label{eq:Iy_Z} \\ 
%     &+\frac{1}{N_y}\sum_{\{z_{j-1}^y\}}(\psi(m_{z_{j-1}}^y+1) - \psi(m_{z_{j-1}}+1)), \nonumber
% \end{align} 
where $\langle\cdot\rangle$ denotes the average over observations, $\psi(\cdot)$ is the digamma function, $N$ is the total number of samples, $N_y$ is the number of samples associated to the outcome $Y=y$, $k$ is the number of neighbors searched for each sample in the space $\{X, \boldsymbol{Z}_{j-1}, V | Y=y\}$, referring to samples in $\{X, \boldsymbol{Z}_{j-1}, V\}$ associated to the outcome $Y=y$, $m_{z_{j-1},v}^y$, $m_{x,z_{j-1}}^y$, and $m_{z_{j-1}}^y$ are the number of samples in $\{\boldsymbol{Z}_{j-1}, V | Y=y\}$, $\{X, \boldsymbol{Z}_{j-1}| Y=y\}$ and $\{\boldsymbol{Z}_{j-1} | Y=y\}$ for the determined distance, respectively, and $m_{x,z_{j-1},v}$, $m_{z_{j-1},v}$, $m_{x,z_{j-1}}$, and $m_{z_{j-1}}$ are those counted considering all samples in $\{X,\boldsymbol{Z}_{j-1},V\}$, $\{\boldsymbol{Z}_{j-1},V\}$, $\{X, \boldsymbol{Z}_{j-1}\}$ and $\{\boldsymbol{Z}_{j-1}\}$.

Given that the set $\boldsymbol{Z}_{min}$ minimizes the information shared between $X$ and $Y$, that determines the unique contribution of the source variable to the class, and that the redundant contribution of the variables in $\boldsymbol{Z}_{min}$ is obtained as the difference between $I(Y;X)$ and $I(Y;X|\boldsymbol{Z}_{min})$, it is necessary to assess first the statistical significance of $I(Y;X)$ to determine if there is any potential unique or redundant contribution. Therefore, in the initial step, the MI term $I(Y;X)$ is computed as
% \begin{align}
% \begin{split}
%     I(Y;X) &= \sum_{y\in A_Y} p(y) I(Y=y;X) \\
%     &= \sum_{y\in A_Y} p(y) \left(\psi(N) - \psi(N_y) + \psi(k) - \frac{1}{N_y}\sum\psi(m_X+1) \right).
% \end{split}
% \end{align}
\begin{equation}
    I(Y;X) = \sum_{y\in A_Y} P(y) I(Y=y;X),
\label{main:I(Y;X)}
\end{equation}
with the specific MI term being
\begin{equation}
\begin{split}
    I(Y=y;X) &= \psi(N) - \psi(N_y)   + \psi(k) - \langle\psi(m_X+1)\rangle;
\end{split}
\end{equation}
here, $k$ is the number of neighbors searched in the space $\{X|Y=y\}$, and $m_X$ is the number of samples counted in the same range for the space $\{X\}$. A surrogate test is performed to
%assess whether the observed value of the measure for variable $X$ is significantly different from what would be expected under the null hypothesis, which assumes that the value arises purely by chance. 
test the null hypothesis of independence between $Y$ and $X$.
To simulate the null distribution, the samples of $X$ are randomly shuffled, and the MI in \eqref{main:I(Y;X)} is recomputed. This procedure is repeated $N_{surr}$ times, producing a distribution of surrogate values. The significance of the original measure is then evaluated by comparing it to the $95^\textit{th}$ percentile of the surrogate distribution. If the observed value exceeds this threshold, the null hypothesis is rejected and the MI is assumed to be greater than 0. Otherwise, this would indicate that the empty set is the final set $\boldsymbol{Z}_{min} = \emptyset$, thus shortening the search procedure.

Similarly, for determined values of $I(Y;X|\boldsymbol{Z}_{j-1}, V_j)$ and $I(Y;X|\boldsymbol{Z}_{j-1})$, a statistical analysis of their difference is performed to test the null hypothesis that the candidate $V_j$ does not alter the conditional information shared by $X$ and $Y$ given $\boldsymbol{Z}_{j-1}$; if the null hypothesis is verified, $V_j$ will not be included in the sets of selected variables $\boldsymbol{Z}_{min}$ or $\boldsymbol{Z}_{max}$ and the search is terminated. Specifically, the significance of $I(Y;X|\boldsymbol{Z}_{j-1}, V_j) - I(Y;X|\boldsymbol{Z}_{j-1})$ is tested for the $\boldsymbol{Z}_{max}$ search, and that of $I(Y;X|\boldsymbol{Z}_{j-1}) - I(Y;X|\boldsymbol{Z}_{j-1}, V_j)$ for the $\boldsymbol{Z}_{min}$ search. With the same null hypothesis, a test is done by randomly permuting the investigated variable $V_j$ $N_{surr}$ times, producing a distribution of surrogate values of the difference measure. If the original measurement exceeds the $95^\textit{th}$ percentile threshold, the difference is deemed significant, and the contributing variable $V_j$ is added to the appropriate set.

Once $\boldsymbol{Z}_{min}$ and $\boldsymbol{Z}_{max}$ are found, all the terms needed to obtain the contributions in \eqref{eq:URS}, i.e., $I(Y;X)$, $I(Y;X|\boldsymbol{Z}_{min})$ and $I(Y;X|\boldsymbol{Z}_{max})$, are computed again considering as neighbor search space $\{X, \boldsymbol{Z}_{min}\cup\boldsymbol{Z}_{max}\}$. Specifically, equations \eqref{eq:IyX_Z} and \eqref{eq:Iy_Z} are employed, replacing $\boldsymbol{Z}_{j-1}$ with $\boldsymbol{Z}_{min}$ and $\boldsymbol{Z}_{max}$. %In the case in which $\{X, \boldsymbol{Z}_{min}\cup\boldsymbol{Z}_{max}\}$ is equivalent to $\{X, \boldsymbol{Z}_{min}\}$ or $\{X, \boldsymbol{Z}_{min}\}$, the formula (8) becomes:

\subsection{Gaussian estimation} \label{main:gaussianestimation}

To validate the accuracy of the estimation approach, we derive the theoretical values of the CMI for a known system. With the assumption that for a given discrete realization $Y=y$, the underlying joint conditioned space $\{X, \boldsymbol{Z}|Y=y\}$ follows a multivariate Gaussian distribution, the joint distribution $p(X,\boldsymbol{Z})$ becomes a Gaussian Mixture Model (GMM), formed by the class weights and conditional probability densities, i.e.,

\begin{equation}
    p(X,\boldsymbol{Z}) = \sum_{\mathcal{A}_Y} P(y)p(X,\boldsymbol{Z}|Y=y).
\end{equation}
To evaluate the CMI values, starting from the definition

\begin{widetext}
\begin{equation}
    I(Y;X|\boldsymbol{Z}) = \sum_{\mathcal{A}_Y}\int_{\mathcal{D}_X}\int_{\mathcal{D}_{\boldsymbol{Z}}} p(y,x,\boldsymbol{z})\log\left(\frac{p(\boldsymbol{z})p(y,x,\boldsymbol{z})}{p(x,\boldsymbol{z})p(y,\boldsymbol{z})}\right)dxd\boldsymbol{z},
\end{equation}
%\end{align}
%we have

%\begin{align}
\begin{equation}
    I(Y;X|\boldsymbol{Z}) =  \sum_{\mathcal{A}_Y}P(y)\int_{\mathcal{D}_X}\int_{\mathcal{D}_{\boldsymbol{Z}}} p(x,\boldsymbol{z}|y)\log\left(\frac{p(\boldsymbol{z})p(x,\boldsymbol{z}|y)}{p(x,\boldsymbol{z})p(\boldsymbol{z}|y)}\right)dxd\boldsymbol{z},
\end{equation}
\end{widetext}
% \begin{align}
% \begin{split}
%     I(Y;X|\boldsymbol{Z}) &= \iiint p(y)p(x,\boldsymbol{z}|y)\log\left(\frac{p(\boldsymbol{z})p(x,\boldsymbol{z}|y)}{p(x,\boldsymbol{z})p(\boldsymbol{z}|y)}\right)dydxd\boldsymbol{z}
%     \\
%     &= \sum_\mathcal{Y}p(y)\iint p(x,\boldsymbol{z}|y)\log\left(\frac{p(\boldsymbol{z})p(x,\boldsymbol{z}|y)}{p(x,\boldsymbol{z})p(\boldsymbol{z}|y)}\right)dxd\boldsymbol{z}
% \end{split}
% \end{align}
with each of the terms, apart from $P(y)$, representing the density for either a multivariate Gaussian distribution or a GMM. The integral represents the specific CMI for a given realization $Y=y$, i.e., $I(Y=y; X | \boldsymbol{Z})$. 

% \begin{align}
% \begin{split}
%     I(&Y=y; X | \boldsymbol{Z}) = \\ &\iint p(x,\boldsymbol{z}|y)\log\left(\frac{p(\boldsymbol{z})p(x,\boldsymbol{z}|y)}{p(x,\boldsymbol{z})p(\boldsymbol{z}|y)}\right)dxd\boldsymbol{z}.
% \end{split}
% \end{align}
% \begin{equation}
%     I_{spec}(Y=y; X | \boldsymbol{Z}) = \iint p(x,\boldsymbol{z}|y)\log\left(\frac{p(\boldsymbol{z})p(x,\boldsymbol{z}|y)}{p(x,\boldsymbol{z})p(\boldsymbol{z}|y)}\right)dxd\boldsymbol{z}.
% \end{equation}
Given the initial distribution assumption constraints, this formulation allows us to simulate complex, multimodal dependencies in the joint space while retaining full knowledge of the underlying generative processes. Since a closed-form expression for the specific CMI is not available in this setting, 
% Since closed-form expressions for the individual specific cross-entropy terms, specifically $H_{X,\boldsymbol{Z}|Y=y}(X,\boldsymbol{Z})$ and $H_{X,\boldsymbol{Z}|Y=y}(\boldsymbol{Z})$, are not available in this setting, 
we resort to high-precision numerical estimation using the Monte Carlo (MC) method \cite{13ab5b5e-0237-33fb-a7a8-6f6e4e0d4e0f}, with the integral approximated as

\begin{align}
\begin{split}
    I(Y=y; &X | \boldsymbol{Z}) =  \hat{\mathbb{E}}_{p(x,\boldsymbol{z}|y)}\left[\log\left(\frac{p(\boldsymbol{z})p(x,\boldsymbol{z}|y)}{p(x,\boldsymbol{z})p(\boldsymbol{z}|y)}\right)\right].
\end{split}
\label{main:MCest}
\end{align}
To obtain this estimate, we draw $10^6$ samples from the conditional distribution space $\{X,\boldsymbol{Z}|Y=y\}$, and for each sample we evaluate the entire log-ratio expression in \eqref{main:MCest}. The MC estimate of $I(Y=y; X | \boldsymbol{Z})$ is then computed as the average of these expressions across all sampled instances.

\section{Validation on synthetic data with mixed Gaussian distributions}

This section presents a series of experiments on synthetically generated data designed in a way such that the underlying generative models are fully specified, allowing to capture the distinct components of information in structured settings. The goals of these experiments are to investigate the true values of the information-theoretic contributions of individual feature variables as a function of the simulation parameters, and to quantify the bias and variance of the proposed estimator relative to the theoretical values of the information measures.
To this end, we first construct a series of simple setups that focus on showcasing each possible type of individual contributions of the feature variables to the target (i.e., unique, redundant, and synergistic), and then design an additional, more complex simulation covering all possible types of interaction.

In detail, we consider a class variable $Y$ and a set of $m$ feature variables $\boldsymbol{X}=\{X_1,\ldots,X_m\}$; the features are assumed to have a multivariate Gaussian distribution when associated with each specific outcome of the class, i.e. $\boldsymbol{X}|Y=y \sim \mathcal{N}(\mu_{\boldsymbol{X}}^{(y)}, \Sigma_{\boldsymbol{X}}^{(y)})$, where $\mu_{\boldsymbol{X}}^{(y)}$ and $\Sigma_{\boldsymbol{X}}^{(y)}$ are the mean vector and the covariance matrix of $\boldsymbol{X}$ evaluated when $Y=y$.
Inspired by the examples in \cite{ontivero2025assessing,wollstadt2023rigorousinformationtheoreticdefinitionredundancy}, we outline a set of simulations inducing behaviors that elicit specific types of information contributions. This is achieved imposing changes in the mean of a feature variable across class states (i.e., imposing different values of $\mu_{X_i}^{(y)}$ at varying $y$) to reproduce relevant contributions, with different levels of high-order interaction determined by the correlation among features (i.e., by the off-diagonal elements of $\Sigma_{\boldsymbol{X}}^{(y)}$), and imposing no changes of mean and variance of a feature across states to reproduce an absence of contribution. In the first three simulations we consider only $m=2$ features and show how unique, synergistic or redundant behaviors can be isolated acting on the simulation parameters, while in the last simulation with $m=6$ we demonstrate how combined changes in the distribution parameters of several features can give rise to coexisting contributions.
The kNN estimation method was applied over simulations lasting $N_{sample}=1000$ samples per realization $Y=y$, considering the typical setting $k=10$ for the number of neighbors \cite{BARA2024380, Xiong2017}; the statistical significance of the selected variables was tested generating $N_{surr}=100$ surrogate sequences. Each simulation was repeated $N_{sim}=100$ times to assess the confidence intervals of each measure, while theoretical values were obtained using $N_{MC}=10^6$ MC samples for each realization $Y=y$.

%Specifically, we expect that a change in the mean of a feature variable across class states represents a unique contribution of that feature in determining the class. However, this contribution may be partially redundant if other features exhibit similar class-dependent changes in the mean, even if those features are uncorrelated. As the correlation between such features increases, the redundancy component grows, while a negative correlation emphasizes their unique contribution. Conversely, a synergistic contribution of the features in determining the class variable can be ascribed to a variation in the correlation of the features across the output of the class variable. Combinations of these properties will give rise to coexistent contributions, which are showcased in the last simulation. 

\subsection{Unique components}
First, we consider a 2-class ($\mathcal{A}_Y=\{y_1,y_2\}$), 2-variable ($\boldsymbol{X}=\{X_1,X_2\}$) setup with the following model parameters:
\begin{align}
\label{simu1}
\begin{split}
    P(y_1) &= P(y_2) = 0.5, \\
    \mu_{\boldsymbol{X}}^{(y_1)} &= [1, 1],\\ \mu_{\boldsymbol{X}}^{(y_2)} &= [-1, 1], \\
    \Sigma_{\boldsymbol{X}}^{(y_1)} &= \Sigma_{\boldsymbol{X}}^{(y_2)} = \begin{bmatrix}
                    1 & 0\\
                    0 & 1
                    \end{bmatrix}.
\end{split}
\end{align}
The class-specific distributions are $X_1|Y=y_1 \sim \mathcal{N}(1,1)$, $X_1|Y=y_2 \sim \mathcal{N}(-1,1)$ and $X_2|Y=y_1 \equiv X_2|Y=y_2 \sim \mathcal{N}(1,1)$, showing that the mean of $X_1$ shifts between the two classes while the distribution of $X_2$ remains the same. The diagonal covariance implies that the two features are uncorrelated.

The results of the analysis are shown in Fig.~\ref{fig:USR}(a), documenting that the predictive information is fully determined by the purely unique contribution of $X_1$, which can be ascribed to the shift in mean between the two class labels; The feature $X_2$, being uncorrelated with $X_1$ and equally distributed within the two classes, displays no relevancy to the class variable $Y$; the uncorrelation between the features explains also the absence of redundant or synergistic effects. The distribution of the information measures estimated over $N_{sim}$ realizations shows the accuracy of the estimates, which exhibit a small negative bias and limited variability.

% \begin{figure}[H]
%     \centering
%     \includegraphics[scale = 0.5]{figures/fig_U.eps}
%     \caption{Unique}
%     \label{fig:U}
% \end{figure}

% \begin{figure}[H]
%     \centering
%     \includegraphics[scale = 0.5]{figures/fig_R.eps}
%     \caption{Redundancy}
%     \label{fig:R}
% \end{figure}

\subsection{Synergistic components}
This simulation reproduces again a 2-class, 2-variable setup, now characterized by the following parameters:
\begin{align}
\label{simu2}
\begin{split}
    P(y_1) &= P(y_2)=0.5, \\
    \mu_X^{(y_1)} &= \mu_X^{(y_2)} = [0, 0], \\
    \Sigma_X^{(y_1)} &= \begin{bmatrix}
                    1 & 0.5\\
                    0.5 & 1
                    \end{bmatrix}, \\
    \Sigma_X^{(y_2)} &= \begin{bmatrix}
                    1 & -0.5\\
                    -0.5 & 1
                    \end{bmatrix},
\end{split}
\end{align}
corresponding to identical standard normal distributions of the two feature variables within the two classes ($X_i|Y=y_j \sim \mathcal{N}(0,1)$, $i,j=1,2$). Moreover, as seen in the covariance matrix, the two features are positively correlated when the class takes the value $y_1$, and anticorrelated when the class takes the value $y_2$.

The results for this setup, reported in Fig.~\ref{fig:USR}(b), document the absence of unique contributions, reflecting the fact that the class-specific mean and variance remained unchanged across the realizations of $Y$. However, their joint interaction, here captured by the class-specific covariance, carries discriminative information about the class, which is encoded in the significant synergistic contribution observed in the absence of redundancy. In this case, synergy arises by the fact that the relationship between the features differs across the realizations of $Y$, so that both $X_1$ and $X_2$ need to be known to predict the class.
%as it is different for each class. In practical terms, this means that marginally the variables appear uninformative (same class-specific mean), but their relationship (i.e., the covariance) differs across the realizations of $Y$, thereby encoding the information about $Y$ synergistically because both need to be known to predict the class. Fig.~\ref{fig:USR}(b) shows that both $X_1$ and $X_2$ show dominantly synergistic contributions.

\subsection{Redundant components}
Another 2-class, 2-variable setup is then considered, featuring the following parameters:
\begin{align}
\label{simu3}
\begin{split}
    P(y_1) &= P(y_2)=0.5, \\
    \mu_X^{(y_1)} &= [1, 1],\\ \mu_X^{(y_2)} &= [-1, -1], \\
    \Sigma_X^{(y_1)} &= \Sigma_X^{(y_2)} = \begin{bmatrix}
                    1 & 0.99\\
                    0.99 & 1
                    \end{bmatrix}.
\end{split}
\end{align}
In this case, the class-specific distributions are the same for the two features, i.e. $X_i|Y=y_1 \sim \mathcal{N}(1,1)$, $X_i|Y=y_2 \sim \mathcal{N}(-1,1)$, $i=1,2$, showing a shift in the mean between the two classes. Moreover, the features are strongly correlated as documented by the covariance value of $0.99$.

\begin{figure*}%[h!]
    \centering
    \includegraphics[width=0.95\linewidth]{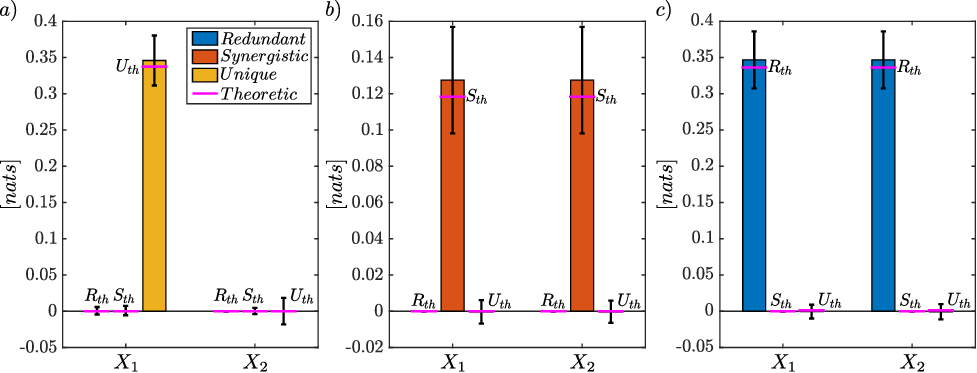}
    \caption{Decomposition into unique, redundant and synergistic components of the maximum information shared between the binary class variable $Y$ and the feature $X_i$ ($i=1,2$) for the simulated systems with parameters given by Eq. \ref{simu1} (a), Eq. \ref{simu2} (b), Eq. \ref{simu3} (c). Barplots represent the mean values (with $\pm 2$ SD) of the various information contributions, for features $X_1$ and $X_2$ showcasing unique (a), synergistic (b), and redundant (c) predictive information; magenta lines represent the theoretical values, computed as in Section~\ref{main:gaussianestimation}.}
    \label{fig:USR}
\end{figure*}

The analysis of this configuration reveals a fully overlapping contribution of the two features to the prediction of the class labels, as seen in Fig.~\ref{fig:USR}(c) where the predictive information is entirely redundant, with zero values for the unique and synergistic components. These results are explained by the fact that the two features share the same class-dependent change in the mean value.

%Due to the requirement in the theoretical estimations, the covariance value of $0.99$ is chosen to have a positive definite covariance matrix. The two variables shared the same class-dependent change in the mean value, revealing redundancy, as seen in Fig.~\ref{fig:USR}(c). 

%We find that the positive increase in correlation (values $>0$) between the variables caused an increase in the redundancy contribution (while lowering the unique component), while the negative increase in correlation (values $<0$) caused an increase in the unique contribution (while lowering the redundancy component), shown in Fig.~\ref{fig:R_expanded}. 

%As seen in Fig.~\ref{fig:USR}(c), both $X_1$ and $X_2$ have dominantly redundant contributions, with minor unique contributions due to the 0.99 correlation.

% \begin{figure}[H]
%     \centering
%     \includegraphics[scale = 0.5]{figures/fig_S.eps}
%     \caption{Synergy}
%     \label{fig:S}
% \end{figure}

%\begin{figure*}[h!]
%    \centering
%    \includegraphics[width=1\linewidth]{figures_new/fig_R_expanded.eps}
%    \caption{Barplots represent the average information contributions for features $X_1$ and $X_2$ showcasing mixed redundant and unique interactions for covariance values of $\text{Cov}(X_1, X_2) = 0$ (in panel \textit{a}), $\text{Cov}(X_1, X_2) = 0.5$ (in panel \textit{b}), and 
%    $\text{Cov}(X_1, X_2) = -0.5$ (in panel \textit{c}), with $\pm2std$ confidence intervals. Magenta lines represent the theoretical values, computed as in Section~\ref{main:gaussianestimation}.}
%    \label{fig:R_expanded}
%\end{figure*}

\subsection{Multiple interactions}

The last simulation, which displays a setup with more complex interactions, including combined contributions of multiple components, is designed to portray the robustness of the proposed estimation strategy, considering a higher-dimensional interaction system. Specifically, a 2-class ($\mathcal{A}_Y=\{y_1,y_2\}$), 6-variable ($\boldsymbol{X}=\{X_1,\ldots,X_6\}$) system is formed with the following model parameters:

\begin{align}
\label{system6}
\begin{split}
    P(y_1) &= P(y_2)=0.5, \\
    \mu_X^{(y_1)} &= [1,0.5,0.5,0,1,0], \\ \mu_X^{(y_2)} &= [-1,-0.5,-0.5,0,-1,0], \\
    \Sigma_X^{(y_1)} &= \begin{bmatrix}
                    1 & 0 & 0 & 0 & 0 & 0\\
                    0 & 1 & 0.25 & 0 & 0 & 0\\
                    0 & 0.25 & 1 & 0 & 0 & 0\\
                    0 & 0 & 0 & 1 & 0.5 & 0\\
                    0 & 0 & 0 & 0.5 & 1 & 0\\
                    0 & 0 & 0 & 0 & 0 & 1\\
                    \end{bmatrix}, \\
    \Sigma_X^{(y_2)} &= \begin{bmatrix}
                    1 & 0 & 0 & 0 & 0 & 0\\
                    0 & 1 & 0.25 & 0 & 0 & 0\\
                    0 & 0.25 & 1 & 0 & 0 & 0\\
                    0 & 0 & 0 & 1 & -0.5 & 0\\
                    0 & 0 & 0 & -0.5 & 1 & 0\\
                    0 & 0 & 0 & 0 & 0 & 1\\
                    \end{bmatrix}.
\end{split}
\end{align}
According to \ref{system6}, the joint distribution of the features display the following characteristics: the feature $X_1$ exhibits different class-specific mean while remaining uncorrelated with the other features; $X_2$ and $X_3$ have the same different class-specific means and additionally display a positive correlation for both realizations of $Y$; while $X_4$ does not exhibit a change in the class-specific mean, $X_5$ does, and they jointly have a class-dependent covariance value; the feature $X_6$ has no class-specific change in the probability distribution and is uncorrelated with the other features. 

\begin{figure}%[tb!]
    \centering
    \includegraphics[width=0.9\linewidth]
    {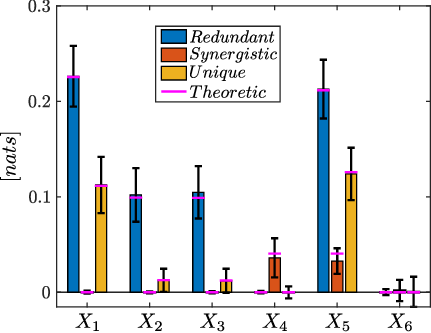}
    \caption{Decomposition into unique, redundant and synergistic components of the maximum information shared between the binary class variable $Y$ and the feature $X_i$ ($i=1,\ldots,6$) for the simulated systems with parameters given by Eq. \ref{system6}. Barplots represent the mean values (with $\pm 2$ SD) of the various information contributions estimated across several realizations, while magenta lines represent the theoretical values, computed as in Section~\ref{main:gaussianestimation}.}
    \label{fig:mixed}
\end{figure}

Fig.~\ref{fig:mixed} shows the results of the kNN estimator applied to the defined system, for which several expected contributions are observed as follows. 
The change in the mean value across class realizations enables the variables $X_1$, $X_2$, $X_3$, and $X_5$ to be predictive for the class, thereby providing unique contribution, with a higher magnitude for $X_1$ and $X_5$ which display the largest mean shift. Moreover, since these feature exhibit similar class-dependent changes in the mean, their contribution is also partially redundant; we note that redundancy arises not only for the correlated variables $X_2$ and $X_3$, but also for $X_1$ and $X_5$ which are not correlated with each other or with $X_2$ and $X_3$. Additionally, a synergistic contribution is detected for $X_4$ and $X_5$, reflecting the change in sign of their correlation across the output values of the class variable $Y$. Finally, $X_6$ does not provide any contribution to $Y$, as expected by the fact that its probability density is not class-dependent and it is uncorrelated with the other features.
Remarkably, the accuracy of the kNN estimation remains high also in this higher-dimensional system, as documented by the very small bias and limited variance of the estimates of unique, redundant and synergistic components computed across $N_{sim}$ realization (barplots and errorbars in Fig.~\ref{fig:mixed}).

\begin{figure*}
    \centering
    \includegraphics[width=0.95\linewidth]
    {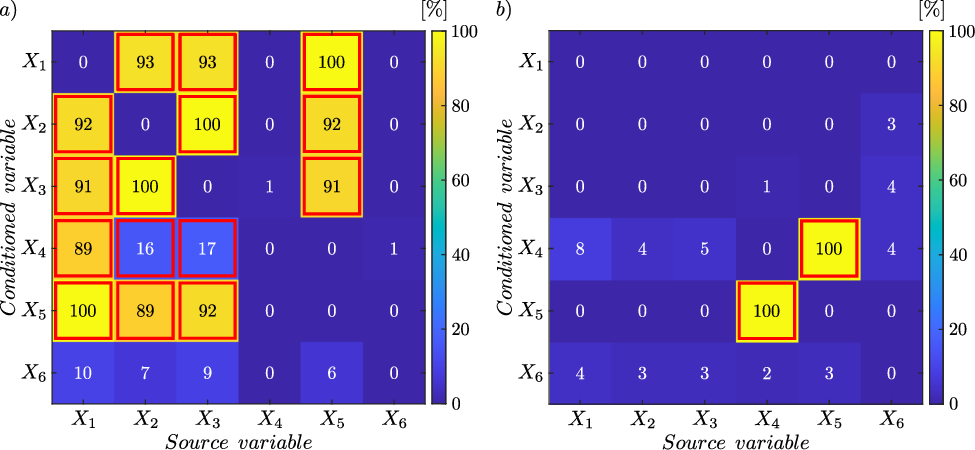}
    \caption{Selection of the features to be included in the conditioning set during the search for redundancy and synergy in the simulated system with parameters given by Eq. \ref{system6}. Heatmaps show, for any source variable (columns), the selection of the conditioning variables (rows; cells with red border: theoretical selection; cell color and number: percentage of selections over $N_{sim}$ realizations) included within the set $\boldsymbol{Z}_{min}$ (\textit{a}) or within the set $\boldsymbol{Z}_{max}$ (\textit{b}). In panel \textit{a}, the source variables $X_1$, $X_2$, $X_3$, and $X_5$ consistently contribute to minimizing the CMI value, with $X_4$ further decreasing it, except when the starting variable is $X_5$. Panel \textit{b} shows that the only synergy-induced CMI increase happens between $X_4$ and $X_5$.}
    \label{fig:Zselection}
\end{figure*}

\begin{figure*}%[tb!]
    \centering
    \includegraphics[width=0.95\linewidth]    {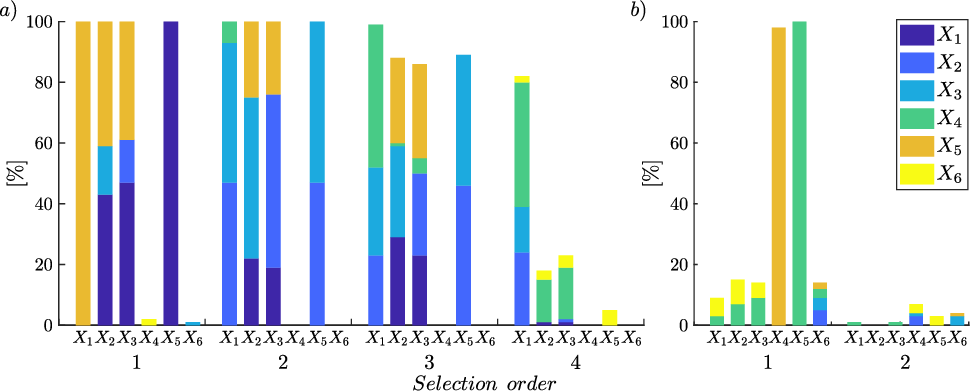}
    \caption{Order and percentage of occurrence of the features selected as conditioning variables for the MI between each source feature and the target class during the search for redundancy (a, inclusion into $\boldsymbol{Z}_{min}$) and synergy (b, inclusion into $\boldsymbol{Z}_{max}$). The height of each bar quantifies the percentage of times over $N_{sim}$ realizations for which a variable was selected for the source source feature $X_i$ ($i=1,\ldots,6$) at each iteration $j\in\{1,2,\cdots\}$; the color information represents which variable was selected.}
    %Variable selection order and occurrence. Bar plots depict the occurrence of selection of the feature variables in determining $\boldsymbol{Z}_{min}$ (in panel \textit{a}) and $\boldsymbol{Z}_{max}$ (in panel \textit{b}), per source variable $X_i$ and iteration $j\in\{1,2,\cdots\}$. The color information represents each of the selected variables.}
    \label{fig:order_selection}
\end{figure*}

Fig.~\ref{fig:Zselection} depicts the selection of the variables included for this simulation into the conditioning sets $\boldsymbol{Z}_{min}$ and $\boldsymbol{Z}_{max}$ during the search for redundant and synergistic contributions to the class label, both theoretical based on the exact CMI values and practical based on the estimated values and the surrogate test for significance.
Moreover, Fig.~\ref{fig:order_selection} shows the order according to which the variables are selected by the sequential procedure for the formation of the sets $\boldsymbol{Z}_{min}$ and $\boldsymbol{Z}_{max}$.

%shows the heatmaps indicating the percentage that the individual variables were selected for conditioning, when determining $\boldsymbol{Z}_{min}$ and $\boldsymbol{Z}_{max}$, for each source variable. Cells with red borders indicate the theoretically estimated connections that contribute to the appropriate sets.
The group of features $X_1$, $X_2$, $X_3$ or $X_5$ are selected into the conditioning set $\boldsymbol{Z}_{min}$ when one of them is taken as source variable (Fig.~\ref{fig:Zselection}a, columns 1,2,3,5), in an order generally prioritizing $X_1$ and $X_5$ followed by $X_2$ and $X_3$ (Fig.~\ref{fig:order_selection}a). This selection is expected given the previously discussed marginal predictive capability of these features, which leads to redundant contributions.
Moreover, $X_4$ is also included theoretically into the set $\boldsymbol{Z}_{min}$ when $X_1$, $X_2$, or $X_3$ are taken as sources; in the practical estimation this selection occurred only in a fraction of realizations ($89\%$,$16\%$ and $17\%$ for $X_1$, $X_2$, and $X_3$, Fig.~\ref{fig:Zselection}a) and at the third or fourth iteration of the searching procedure (Fig.~\ref{fig:order_selection}a). The inclusion of $X_4$ is more nuanced, as it exhibits a synergistic interaction with $X_5$, and its selection along with $X_5$ further decreases the CMI.
%Starting with feature $X_1$, the features selected for $\boldsymbol{Z}_{min}$ are $X_2, \dots,X_5$ (Fig.~\ref{fig:Zselection}(a), first column). Here, the selection of $X_2$, $X_3$, and $X_5$ is expected, given their previously discussed marginal predictive capabilities. However, the inclusion of $X_4$ is more nuanced, as it exhibits a synergistic interaction with $X_5$, and its selection along with $X_5$ further decreases the CMI. A similar reasoning can be applied when starting with $X_2$ or $X_3$.
%In contrast, when starting with $X_5$, the selected features are limited to $X_1$, $X_2$, and $X_3$, as including $X_4$ would increase the CMI, due to its synergistic interaction with $X_5$. 

With regard to the procedure of maximization of the CMI, the only significant selections involve $X_4$ and $X_5$, which are always reciprocally included into $\boldsymbol{Z}_{max}$ at the first iteration (Fig.~\ref{fig:Zselection}b, Fig.~\ref{fig:order_selection}b). This selection is meaningful as it relates to the synergistic contribution to the class labels expected for the features $X_4$ and $X_5$. Only minor spurious inclusions, not expected theoretically and related to false positive selections, are observed in the other cases.

%%selection order of individual (colored) variables within the selected $X_i$ source to form $\boldsymbol{Z}_{min}$.

\section{Application to non-Gaussian data}

To investigate the behavior of the kNN CMI estimator when dealing with distributions for which the theoretical values of the information components cannot be computed, we design two validation scenarios: one using synthetic data and the other using real-world data. In the synthetic case, we interpret the results according to the expected interaction effects based on the experimental setup, while in the real-world case, we validate the findings against known analyses reported in the literature.

\subsection{Synthetic data} \label{synthetic}

In this setup, we simulate from four independent uniform distributions, $X_i \sim \mathcal{U}(-1,1)$, $i \in \{1,\dots,4\}$. The class discrete variable $Y$ is formed as $Y = \Theta(X_1 \cdot X_2) + \Theta(X_3) + \Theta(X_4)$, where $\Theta(\cdot)$ denotes the Heaviside function, resulting in a class alphabet $A_Y=\{0,1,2,3\}$. Across $N_{sim}=100$ simulations, the proposed estimator receives as inputs the class variable $Y$ and the set of feature variables $X_1,...,X_4$, along with two additional features $X_5$ and $X_5$ defined as $X_5 = X_4 +M$ (a noisy version of $X_4$ with $M \sim \mathcal{N}(0,1)$), and as an independent uniform variable $X_6 \sim \mathcal{U}(-1,1)$. Based on this setup, we expect a synergistic effect between $X_1$ and $X_2$, as they interact in a product-wise manner (like in the XOR logic-gate), while $X_3$ and $X_4$ are expected to provide unique contributions. Moreover, due to the additive nature of the formulation of the class variable $Y$, we anticipate some amount of synergistic interaction among these four variables as well. The variable $X_5$, being a noisy version of $X_4$, is expected to share with it a certain amount of redundant predictive information. Finally, $X_6$, being independent from all other variables, should show no contributions to the target class.
To evaluate the sensitivity of the measures to sample size, the analysis was repeated using $N_{sample}=1000$ and $N_{sample}=300$ as sample size.
%we repeat the simulation with $N_{samples}=300$ and compare the results.
% The GMM parameters are estimated from the data by computing the class-conditional means $\mu_y$ and covariance matrices $\Sigma_y$ for each realization $Y=y$. \textcolor{red}{The covariance matrix is estimated using normalization by $N_{samples}-1$, $\hat{\Sigma_y} = \frac{1}{N_{samples}-1}W_c^TW_c$, where $W$ denotes samples from a multivariate Gaussian distribution (rows as samples), and $W_c$ is the mean centered version of $W$.}

% In the second setup, the same initial inputs are used, but the target variable is defined as $Y = \Theta(X_1 \cdot X_2 + X_3 + X_4)$.

\begin{figure*}%[t]
    \centering
    \includegraphics[width=0.9\linewidth]
    {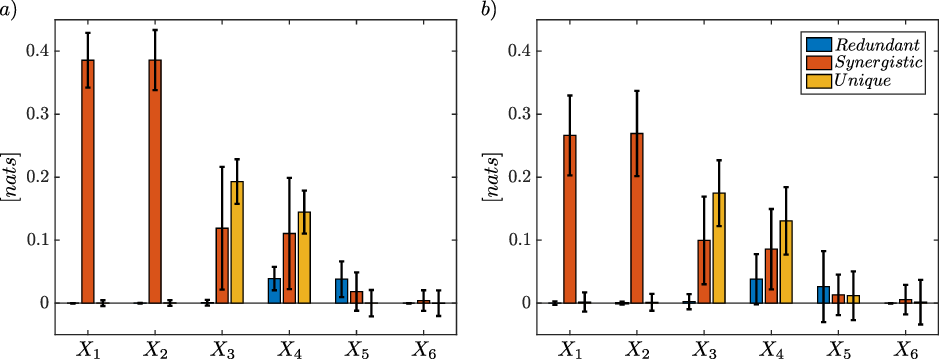}
    \caption{Decomposition into unique, redundant and synergistic components of the maximum information shared between the 4-class variable $Y$ and each of the six source feature variables simulated as in Sect. 4.1. 
    Barplots represent the mean value ($\pm$2 SD) of the various information contributions estimated across several realizations lasting $N_{sample}=1000$ (\textit{a}) and $N_{sample}=300$ (\textit{b}).}
    \label{fig:expSyn}
\end{figure*}

Results in Fig.~\ref{fig:expSyn} confirm the expected synergistic interaction of $X_1$ and $X_2$, as well as the unique contributions of $X_3$ and $X_4$. Additionally, we correctly identify redundancy in $X_4$ since $X_5$ represents a noisy version of it. Finally, $X_6$, which is independent of the other variables, shows no contributions concerning the class. We additionally confirm the expectation of the synergistic effect caused by the additive effect of $X_1,X_2$ with $X_3$ and $X_4$. Notably, the observed synergistic effect has lower values when using a smaller sample size. 
% that both estimators detect similar levels of unique and redundant information contributions. However, a noticeable difference appears when comparing the synergistic influence, with the Gaussian model showing lower values for this type of interaction.  \textcolor{red}{This is potentially due to the inability of the correlation values (captured within the class-covariance matrices) to accurately represent higher-order interactions, which leads to an underestimation of synergy or that the initial assumption for a jointly distributed Gaussian for a class realization is wrong, as the true variable interaction can follow a different, non-Gaussian, distribution. A potential solution is to explicitly include candidate interaction terms, analogous to creating interaction features in a linear or logistic regression framework.}

% \begin{figure}[H]
%     \centering
%     \begin{subfigure}[b]{0.45\columnwidth}
%     \centering
%     \includegraphics[width=1\columnwidth]{figures/figSynthKNN1.eps}
%     \end{subfigure} \vspace{0.001cm}
% ~
%     \begin{subfigure}[b]{0.45\columnwidth}
%     \centering
%     \includegraphics[width=1\columnwidth]{figures/figSynthGMM1.eps}
%     \end{subfigure} \vspace{0.001cm}

%     \caption{Setup 2. Barplots represent the contributions obtained with the kNN-based estimation approach (in panel \textit{a}) and with the Gaussian approximations (in panel \textit{b}). The $95\%$ confidence intervals are shown for all contributions.}
%     \label{fig:exp2}
% \end{figure}

\subsection{Real-world data}

For this analysis, we utilize a subset of The Cancer Genome Atlas - Breast Invasive Carcinoma (TCGA-BRCA) RNA sequencing dataset \cite{Tomczak2015} containing gene expression measurements from breast cancer tissues. The data, obtained from the Marginal Contribution Feature Importance (MCI) \cite{catav2020marginalcontributionfeatureimportance} (\href{https://github.com/TAU-MLwell/Marginal-Contribution-Feature-Importance/tree/main/BRCA_dataset}{GitHub page}), comprises 572 samples characterized by the expressions levels of 50 genes, spanning four molecular subtypes: 304 Luminal A (LumA), 114 Luminal B (LumB), 105 Basal, and 49 HER2-enriched. The approach introduced in this work is used to investigate the contribution of gene expressions in determining the molecular subtype, considered as class outputs in the analysis. To estimate the confidence intervals of the informational contribution of each feature, we generate a bootstrapped dataset with 572 samples and $N_{sim}=100$ sampling repetitions, while ensuring the same class counts. Using the kNN estimator, we fix the number of neighbors at $k=10$, and employ $N_{surr}=100$ surrogate samples for the selection criteria.

\begin{figure*}%[tb!]
    \centering
    \includegraphics[width=0.95\linewidth]{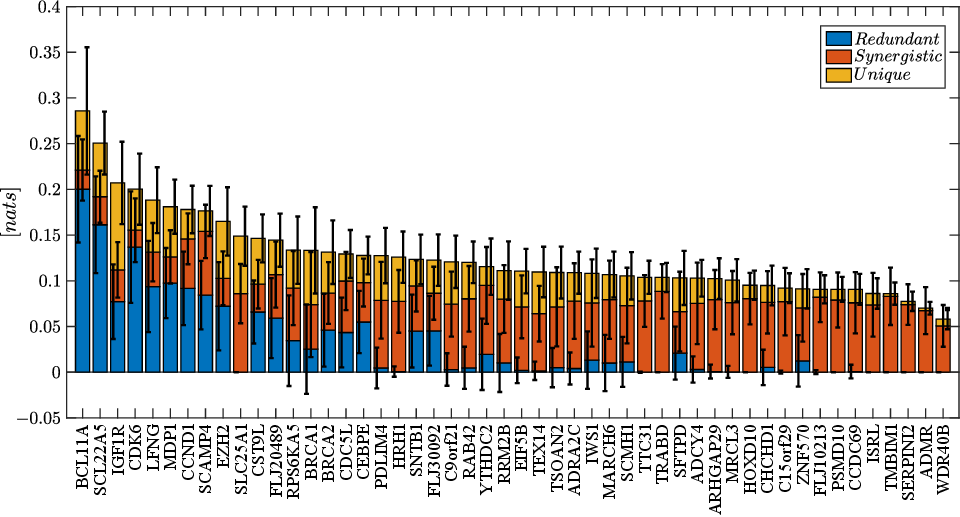}
    \caption{TCGA-BRCA feature analysis results. Barplots represent the contributions obtained with the kNN-based estimation approach for each gene, shown according to a decreasing CMI. The bar and confidence intervals represent the mean $\pm$2 SD across 100 bootstrap samples obtained for the unique, synergistic, and redundant contributions for each gene.}
    \label{fig:realData}
\end{figure*}

The results shown in Fig.~\ref{fig:realData} display the ranked top contributors according to the average CMI estimated between the target class and each observed feature, conditioned on the subset of features that maximizes this measure, $I(Y;X|\boldsymbol{Z}_{max})$. According to \cite{westphal2024partialinformationdecompositiondata, janssen2024ultramarginalfeatureimportancelearning}, the 10 genes: BCL11A, SLC22A5, CDK6, IGF1R, LFNG, CCND1, EZH2, BRCA1, BRCA2, and TEX14, are known breast cancer-associated markers; in our analysis, most of these genes exhibit high CMI values based on their mean contributions. The findings, based on the selected set of features, are consistent with previous, similar works in \cite{janssen2024ultramarginalfeatureimportancelearning, covert2020understandingglobalfeaturecontributions, catav2020marginalcontributionfeatureimportance}, that identify the top contributing set of genes. However, as also noted by  Catav \textit{et al.} \cite{catav2020marginalcontributionfeatureimportance}, we observe substantial contributions from MDP1, SCAMP4, SLC25A1, and CST9L genes, some of which are not linked to breast cancer based on the literature. SLC25A1, however, has been reported by Catalina-Rodriguez \textit{et al.} \cite{CatalinaRodriguez2012} to exhibit elevated expression in breast cancer tissues, suggesting a potential relevance to subtype differentiation. 
% Albeit not comparable in this setting, we note the low influence of BRCA1, BRCA2, and TEX14 despite being one of the prognostic markers.

% \begin{figure}[H]
%     \centering
%     \begin{subfigure}[b]{1\columnwidth}
%     \centering
%     \includegraphics[width=1\columnwidth]{figures/fig_BRCA_knn.eps}
%     \end{subfigure} 

%     \begin{subfigure}[b]{1\columnwidth}
%     \centering
%     \includegraphics[width=1\columnwidth]{figures/fig_BRCA_gauss.eps}
%     \end{subfigure} 

%     \caption{TCGA-BRCA dataset results. Barplots represent the contributions obtained with the kNN-based estimation approach (in panel \textit{a}) and with the Gaussian approximations (in panel \textit{b}). The $95\%$ confidence interval is shown for the unique (U), synergistic (S), and redundant (R) contribution for each gene, from left to right.}
%     \label{fig:BRCA}
% \end{figure}
%\textcolor{red}{We could follow how much of a reduction/increase happens for these conclusions... Or the specific CMI...}
Considering the 10 selected genes, the proposed decomposition allows for the observation of component-wise interactions, noting that the majority of the significant contribution comes from the redundancy components. A similar study by Westphal \textit{et al.} \cite{westphal2024partialinformationdecompositiondata} performs a PID-like analysis to determine synergistic and redundant contributions in addition to the MI that each feature has with the target, $I(Y;X_i)$. However, while their results indicate lower contribution of synergy and redundancy, we note that the nature of these components is different. Specifically, they define feature-wise synergy and redundancy (FWS and FWR) through the concept of interaction information. To obtain FWS they maximize $I(Y;X_i;\boldsymbol{Z}^{ms})$, where $\boldsymbol{Z}^{ms}$ is the subset of features for maximum synergy, while the FWR is determined as $\text{FWS} - I(Y;X_i;\boldsymbol{Z}_{ \setminus X_i})$, where $\boldsymbol{Z}_{\setminus X_i}$ is the subset of features without $X_i$.

For our approach, the redundancy measure for a given gene indicates shared information when the inclusion of another gene results in a reduced CMI. Analyzing the selection order (not shown in the work), we observe shared redundancy between BCL11A and SLC22A5, which is expected given their subtype-specific roles as BCL11A is associated with the basal-like phenotype \cite{Ktnik2023}, whereas SLC22A5 is linked to luminal phenotypes and is known to be an estrogen receptor (ER) regulated gene \cite{Orrantia-Borunda2022-zl, Wang2012}. Furthermore, LFNG and BCL11A exhibit redundant information, likely due to their shared association with the basal-like subtype \cite{Xu2012}. 

\section{Conclusions}

In this work, we presented a method to assess feature importance accounting for high-order interactions among features, designed by fully leveraging an information-theoretic approach based on CMI, estimated using the kNN algorithm, with a focus on classification problems involving continuous features. This approach enables the disentanglement of individual feature contributions into unique, synergistic, and redundant components in a model-independent manner.

Through evaluation on synthetic datasets with known Gaussian distributions, we validated the accuracy of the proposed method against theoretical expectations. Additional experiments on non-Gaussian data, including a real-world breast cancer gene expression dataset, demonstrated that the estimator reliably captured the expected interaction patterns based on the experimental setup or established findings in the literature.

Compared to PID, the proposed CMI-based Hi-Fi extension offers an alternative perspective for analyzing high-order feature interactions. While PID typically focuses on decomposing the total mutual information into predefined atomic components requiring a redundancy measure, whose definition is debated in literature, our approach provides a more flexible, observation-driven decomposition based on traditional information-theoretic measures. Additionally, our method naturally accommodates direct CMI estimation without the need to exhaustively compute all subset combinations, offering scalability and practical relevance for larger systems.

The proposed framework offers potential applications in machine learning, particularly as a feature selection criterion by prioritizing features with predominantly unique or synergistic contributions. Similarly, it can assist in the development of models that better capture the interaction patterns by explicitly accounting for the information structure of the inputs.
Nevertheless, further investigation is necessary to assess the estimator's performance in high-dimensional, highly interactive systems, where the curse of dimensionality may introduce challenges for CMI estimation, as well as to complete a more formal analysis comparing our approach with others proposed in the literature.

\section*{Acknowledgement}
This research was co-funded by the Italian Complementary National Plan PNC-I.1 "Research initiatives for innovative technologies and pathways in the health and welfare sector” D.D. 931 of 06/06/2022, "DARE - DigitAl lifelong pRevEntion" initiative, code PNC0000002, CUP: B53C22006460001, and by the project "HONEST - High-Order Dynamical Networks in Computational Neuroscience and Physiology: an Information-Theoretic Framework”, Italian Ministry of University and Research (funded by MUR, PRIN 2022, code 2022YMHNPY, CUP: B53D23003020006).
I.L. and N.J. are supported by the Ministry of Science, Technological Development and Innovation (Contract No. 451-03-137/2025-03/200156) and the Faculty of Technical Sciences, University of Novi Sad, through the project “Scientific and Artistic Research Work of Researchers in Teaching and Associate Positions at the Faculty of Technical Sciences, University of Novi Sad 2025” (No. 01-50/295).

\bibliography{main_arXiv.bib}

%\nocite{*}
%\bibliography{main.bib}

\end{document}